\documentclass[conference]{IEEEtran}
\IEEEoverridecommandlockouts
\usepackage{cite}
\usepackage{amsmath,amssymb,amsfonts}
\usepackage{algorithmic}
\usepackage{graphicx}
\usepackage{subfig}
\usepackage{textcomp}
\usepackage{xcolor}
\def\BibTeX{{\rm B\kern-.05em{\sc i\kern-.025em b}\kern-.08em
    T\kern-.1667em\lower.7ex\hbox{E}\kern-.125emX}}
\begin{document}

\title{Knowledge Distillation via  Weighted Ensemble of Teaching Assistants\\}

\author{\IEEEauthorblockN{Durga Prasad Ganta}
\IEEEauthorblockA{\textit{Department of Computer Science} \\
\textit{Texas Tech University}\\
Lubbock, Texas \\
durgaprasad.ganta@ttu.edu}

\and
\IEEEauthorblockN{Himel Das Gupta}
\IEEEauthorblockA{\textit{Department of Computer Science} \\
\textit{Texas Tech University}\\
Lubbock, Texas \\
Himel.Das@ttu.edu}

\and
\IEEEauthorblockN{Victor S. Sheng}
\IEEEauthorblockA{\textit{Department of Computer Science} \\
\textit{Texas Tech University}\\
Lubbock, Texas \\
 victor.sheng@ttu.edu}
}

\maketitle

\begin{abstract}
Knowledge distillation in machine learning is the process of transferring knowledge from a large model called teacher to a smaller model called student. Knowledge distillation is one of the techniques to compress the large network (teacher) to a smaller network (student) that can be deployed in small devices such as mobile phones. When the network size gap between the teacher and student increases, the performance of the student network decreases. To solve this problem, an intermediate model is employed between the teacher model and the student model known as the teaching assistant model, which in turn bridges the gap between the teacher and the student. In this research, we have shown that using multiple teaching assistant models, the student model (the smaller model) can be further improved. We combined these multiple teaching assistant model using weighted ensemble learning where we have used a differential evaluation optimization algorithm to generate the weight values. 
\end{abstract}


\section{Introduction}
In the past 10 years, the best-performing artificial-intelligence systems such as the speech recognition on mobile devices or Google’s automatic translator have resulted from an advanced techniques which uses deep neural networks and models that uses deep neural networks are called deep learning models \cite{b1}. Deep neural networks can be constructed and trained with more and more data. With the adequate amount of data their performance continues to increase, which is different from other machine learning techniques that reach a plateau in performance. It is a known fact that performance and accuracy of a deep neural network model increases by employing more layers and parameters \cite{b4}. It is now possible to train very deep models with thousands of layers on efficient GPU or TPU clusters using a variety of new techniques, such as residual connections and batch normalization. For example, training a ResNet model on a famous image recognition benchmark with millions of images takes less than ten minutes. A effective BERT \cite{b7} model for language understanding can be trained in less than one and a half hours. Large-scale deep models have had a lot of success, but their high computational complexity and vast storage requirements make them difficult to use in real-time applications, particularly on low-resource devices like mobile phones and autonomous vehicles. To address this problem, model compression techniques were introduced. One such technique is knowledge distillation \cite{b2} which is transferring knowledge from a large model to a smaller model without loss of validity. As smaller neural network models are less expensive compared to large models, they can be deployed on less powerful hardware like a smartphone.
\begin{figure}[htp]
    \centering
    \includegraphics[scale=0.4]{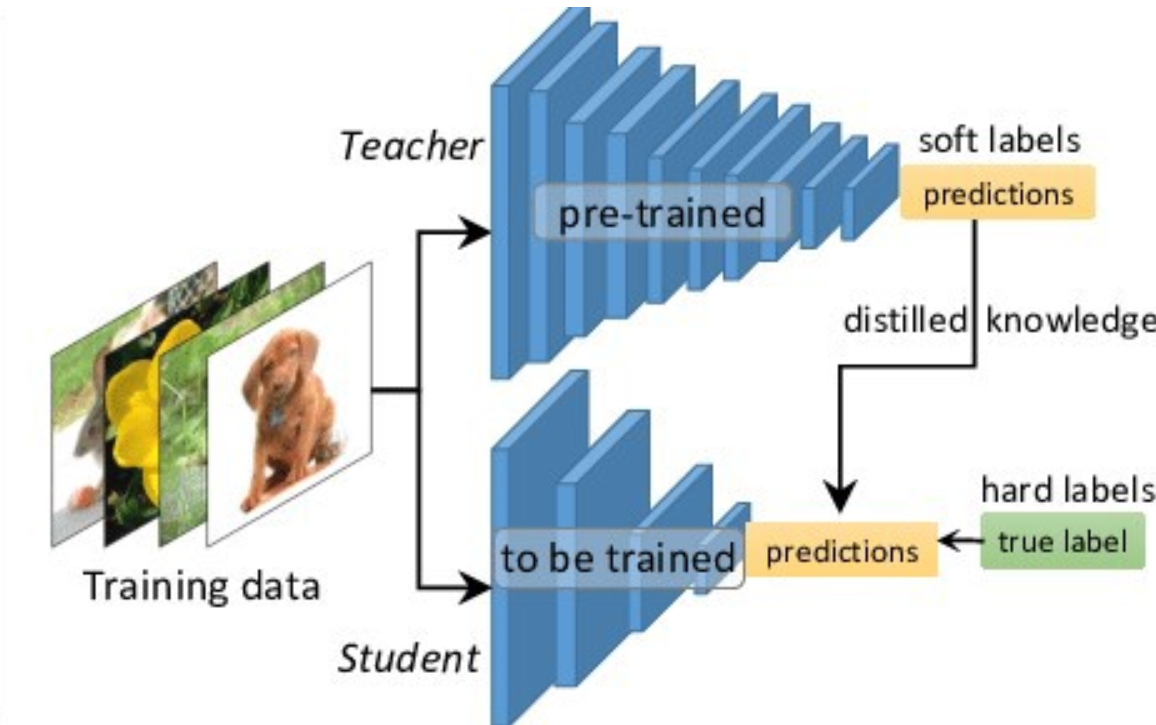}
    \caption{Knowledge distillation}
    \label{fig:n=3}
\end{figure}
\par
However, one of the problems of knowledge distillation is the gap between teacher and student. It has been proved that teacher model with more parameters and good accuracy cannot improve the performance of a student model all the time. Hence, a new method is proposed to improve the performance of student model by employing an intermediate model between a large teacher model and a smaller student model \cite{b5} shown in fig 1. The following are observed,
\begin{figure}[htp]
    \centering
    \includegraphics[scale=0.3]{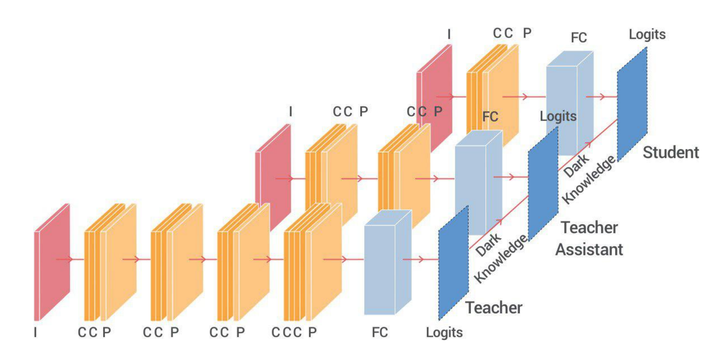}
    \caption{Teaching assistant model}
    \label{fig:n=3}
\end{figure}
\begin{enumerate}
    {\item It is very important to reduce the gap between student and teacher models.
    \item The accuracy of a student can be improved even though the teacher is very large by employing a teaching assistant model.
    \item This framework can be further extended by using a chain of teaching assistant models.}
\end{enumerate} 
\par
Ensemble learning \cite{b20} is one of the most powerful techniques introduced to improve the performance of deep learning models. In comparison with individual model, prediction accuracy over the test set can be significantly boosted by simply averaging the output of a few independently trained neural networks over the same training data set. The size of an ensemble model is not constrained by training because the sub-networks can be trained independently, and their outputs can be computed in parallel.
\par
In this paper, a new method is proposed in which instead of a single teaching assistant model in between teacher and student model, multiple teaching assistants are used. The motivation behind this implementation is the real life scenario like a regular classroom. In a classroom, we often see multiple teacher assistants(TA) helping the students in multiple sectors. There are teaching assistants for labs, grading as well for lecturing in the class. Together they create a spectacular environment that helps the student to amplify their knowledge. To be specify, each teaching assistant model  have different architecture from one another. These teaching assistant models \cite{b19} gets their knowledge distilled from a single large teacher model. Later, predictions of these multiple teaching assistant models are combined using a method of ensemble learning and obtained knowledge is used to train the student model. The contribution of the paper is pointed below,

\begin{itemize}
    \item We have implemented a combination of multiple teaching assistant models in knowledge distillation.
    \item We have used an ensemble learning method to combine the multiple teaching assistant models.
    \item We have used two ensemble learning methods, simple averaging and weighted averaging and compared the result between both.
    \item In the weighted averaging algorithm, we have used differential evaluation optimization method to calculate the weights for each teacher.
\end{itemize}

The following part of this paper is organized in this way. In section II, we will discuss the related works done in this field. In the section III, we will describe our proposed method regarding how we model our experiment and what is the contribution. In section IV, we will discuss our experiment setup. The obtained results and the discussion on those results are shows in section V and VI. We will finish the paper with conclusion and future works in section VII.
\section{Related Work}

In a broader scheme of machine learning, knowledge distillation is basically one form of model compression. There are also some other methods that the researchers have proposed over the years. Now we are going to discuss some of those methods.
\subsection{Model Compression:}

    As the main goal is to have the lighter model and have better accuracy \cite{b6}, there have been researches to achieve this goal. One of them is to reduce the connections based on weight magnitudes. The reduced network is fine-tuned on the same dataset to retain its accuracy. Some of the well known techniques have been described below.
    \begin{enumerate}
        \item \emph{Parameter pruning and sharing:} These methods focus on removing parameters which are not crucial to the model performance from deep neural networks without any significant effect on the performance. This category is further classified into model quantization, model binarization, structural matrices  and parameter sharing.
        \item \emph{Low-rank factorization:} These approaches use matrix and tensor decomposition to find redundant parameters in deep neural networks. This technique factorizes the original matrix into lower rank matrices while preserving latent structures and addressing the issue of sparseness.
        \item \emph{Transferred compact convolutional filters:} Compact convolutional filters can minimize associated computational costs directly. This compression method's main concept is to substitute over-parametric filters with compact filters in order to achieve overall speed while retaining comparable accuracy.
        \item \emph{Knowledge Distillation:} Knowledge distillation is a model compression technique that involves training a small (student) model to match a large pre-trained (teacher) model. By minimizing a loss function, knowledge is transferred from the teacher model to the student model, with the goal of matching softened teacher logits and ground-truth labels. By using a "temperature" scaling function in the softmax, the logits are softened, essentially smoothing out the probability distribution and revealing the teacher's inter-class relationships. The following section will discuss in detail about knowledge distillation.
    \end{enumerate}
    
    \subsection{Knowledge distillation}
    Originally proposed by Bucila, Caruana, and Niculescu-Mizil (2006) \cite{b6} and popularized by Hinton, Vinyals, and Dean (2015) \cite{b2} knowledge distillation compress the knowledge of a large and computational expensive model (often an ensemble of neural networks) to a single computational efficient neural network. The idea of knowledge distillation is to train the small model, the student, on a transfer set with soft targets provided by the large model, the teacher. Knowledge distillation has been commonly used in a number of learning tasks since then. Modeling knowledge transfer between teacher and student has also been done using adversarial methods. Using several teachers was still a good way to improve robustness. Some studies also proposed deep mutual learning which allows an ensemble of student models to learn collaboratively and teach each other during training.
    \par The main idea of using knowledge distillation is that student network (S) to be trained not only using the true labels information but also observation of how the teacher (T) works with the unseen data provided. Because the teacher model has more more generalization power, the idea is to train the student model is such way that it can mimic the behaviour of that generalization. The teacher network is complex in size being deeper and wider.
    \par
    Let $a_t$ and $a_s$ be the logits (the inputs to the final softmax) of the teacher and student network, respectively. In classic supervised learning, the mismatch between output of student network softmax($a_s$) and the ground-truth label $y_r$ is usually penalized using cross-entropy loss and is given as,
    \begin{center}
        $\mathcal{L}_{SL}$=$\mathcal{H}$(softmax($a_s$),$y_r$)
    \end{center}
    In knowledge distillation one also tries to match the softened outputs of teacher $y_t$=softmax($a_t$) and student $y_s$=softmax($a_s$) via Kullback–Leibler divergence loss,
    \begin{center}
        $\mathcal{L}_{KD}$= $\tau^2KL$($y_s$,$y_t$)
    \end{center}  by using a temperature parameter $\tau$ which has an additional control on softening of signal arising from the output of the teacher network. The student network is then trained under the following loss equation which used KD loss and cross-entropy loss,
    \begin{center}
    $\mathcal{L}_{student}$=(1-$\lambda$) $\mathcal{L}_{SL}$+$\mathcal{L}_{KD}$   \end{center} where $\lambda$ is the parameter used to trade-off between these two losses. This method is used for knowledge distillation from teacher and student models.

    \subsection{Distillation Schemes}
     According to how the teacher and student model are trained i.e, whether teacher and student models are trained simultaneously or not, the distillation schemes are divided into following three categories, they are \textbf{offline distillation, online distillation} and \textbf{self-distillation}:\\
    
        \subsubsection{Offline Distribution} Most of the knowledge distillation methods works offline. For the knowledge distillation introduced by Hinton in 2015 \cite{b2}, offline distillation is used which means the knowledge is transferred from a pre-trained teacher model into a student model. The offline distillation process have two stages, they are 1) firstly, training the large teacher model  on a set of training samples before distillation; and 2) then the knowledge is extracted from the  teacher model in the forms of logits or the intermediate features, which are then used to guide the training of the student model during distillation.
        \begin{figure}[htp]
        \centering
        \includegraphics[scale=0.6]{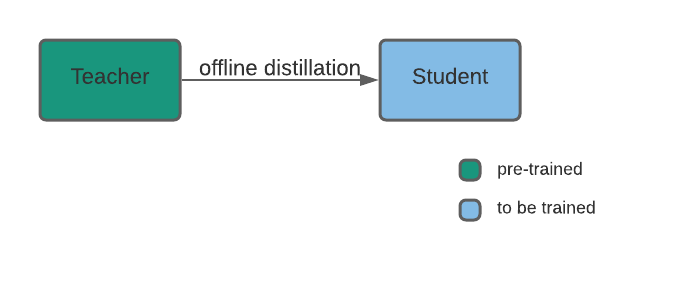}
        \caption{Offline distillation}
        \label{fig:n=3}
        \end{figure}
        \par 
        In the first stage of offline distillation, it is generally assumed that teacher model is pre-defined. Because of that few amount of attention is paid to the teacher model system and its relationship with the student model. As a result, offline distillation method primarily concentrate on enhancing various aspects of information transfer, such as knowledge design and loss functions \cite{b17} for matching features or matching distributions. The main advantage of offline distillation methods is that they are quick and straightforward to use. The teacher model, for example, could include a set of models trained with various software packages and probably on different machines. The knowledge can be extracted and stored in a cache. One-way knowledge transfer and a two-phase training method are popular in offline distillation methods. However, the complex high-capacity teacher model, which requires a significant amount of training time, cannot be avoided, while the student model's offline distillation training is normally effective when guided by the teacher model.
         \subsubsection{Online Distillation} Despite the fact that offline distillation methods are simple and reliable, some issues in the field are gaining more attention from the research community. To overcome the limitations of offline distillation, online distillation \cite{b8} is proposed to boost the student model's performance even more, particularly when a large-capacity high-performance teacher model is not accessible. Both the teacher and student models are updated concurrently in online distillation, and the entire knowledge distillation framework is trainable from beginning to end. A multi-branch architecture was proposed \cite{b11} to minimize computational costs, in which each branch represents a student model and multiple branches share the same backbone network.
        \begin{figure}[htp]
        \centering
        \includegraphics[scale=0.6]{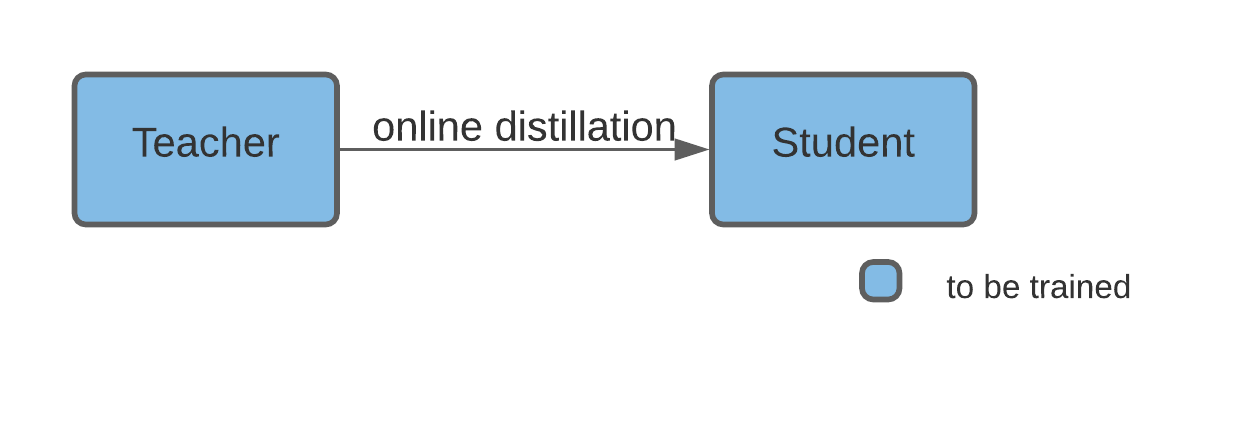}
        \caption{Online distillation}
        \label{fig:n=3}
        \end{figure}
        \par Recently, several online knowledge distillation methods have been proposed \cite{b8,b9, b10}. It's a novel way to make multiple neural networks collaborate in deep mutual learning. A multi-branch architecture was proposed \cite{b11} to minimize the computational costs, in which each branch represents a student model and multiple branches share the same backbone network. Instead of  using the ensemble of logits, a new strategy is introduced \cite{b12} where a feature fusion module has been implemented to construct the teacher classifier. In another strategy \cite{b13},  convolution layer is replaced with cheap convolution operations to form the student model. Another variant of online distillation called co-distillation is employed to train large-scale distributed neural network. Co-distillation is a process in which multiple models are trained in parallel with the same architectures. Online distillation is a one-phase end-to-end training scheme that uses parallel computing to maximize performance. Established online methods (e.g., mutual learning) often struggle to answer the high-capacity teacher in online settings, making it an intriguing subject to further investigate the relationships between teacher-student model in online settings.
         \subsubsection{Self-Distillation:} The same networks are used for the teacher and the student models in self-distillation. This can be regarded as a special case of online distillation. Specifically,  a new self-distillation method is proposed \cite{b14} in which knowledge is distilled from the deeper sections of the network into its shallow sections.  The network in a new self-attention distillation method proposed for lane detection utilizes the attention maps of its own layers as distillation targets for its lower layers. 
        \begin{figure}[htp]
        \centering
        \includegraphics[scale=0.6]{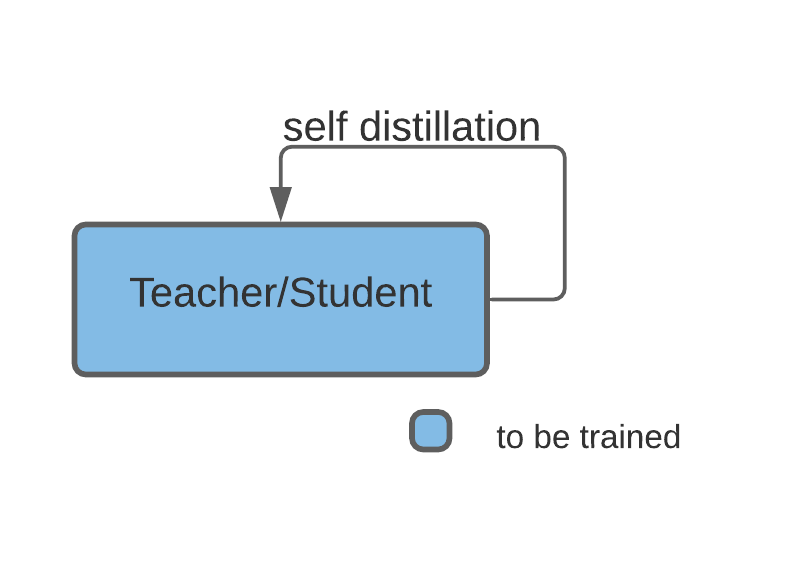}
        \caption{Self-distillation}
        \label{fig:n=3}
        \end{figure}
        \par 
        A special variant of self-distillation called Snapshot distillation in which knowledge in the earlier epochs of the network (teacher) is transferred into its later epochs (student) to support a supervised training process within the same network. To reduce inference time even further through early exit, a new distillation-based training scheme is proposed, in which the early exit layer tries to mimic the performance of the later exit layer during training. Furthermore, interesting self-distillations methods are proposed, one of which is teacher-free knowledge distillation \cite{b15} methods based on the analysis of label smoothing regularization. Instead of conventional soft probabilities, a new self-knowledge distillation process is used, in which the self-knowledge is composed of predicted probabilities. The feature representations of the training model define these expected probabilities. They show how data in function embedding space is identical.
        \par Furthermore, offline, online, and self distillation can all be intuitively understood through the perspective of human teacher-student learning. Offline distillation refers to when a professional teacher teaches a student something to make the student gain knowledge about it; online distillation refers to when both the teacher and the student study together; and self-distillation refers to when the student learns knowledge on their own. Furthermore, similar to how humans learn, these three types of distillation can be combined to complement each other depending on individual advantages. For example, both self-distillation and online distillation are properly integrated via the multiple knowledge transfer framework \cite{b16}.
        We have further discussed the mathematical aspect of the distillation theory in the next section.

     \subsection{Ensemble modeling}
    Ensemble learning, is a method for predicting an outcome that employs a number of different base models. The objective of using ensemble models is to reduce the prediction's generalization error. In order to make a prediction, the method looks to the wisdom of crowds. Despite the fact that the ensemble model contains multiple base models, it behaves and performs as a single model. Main types of ensemble methods are: 
    \begin{enumerate}
        \item \textit{Bagging:} The term "bagging" refers to the process of aggregating data using bootstraps. It is primarily used in classification and regression. It improves model accuracy by employing decision trees, which minimize variance to a large extent. Reduced variance improves precision, which eliminates over-fitting, which is a problem for many predictive models.
        \item \textit{Boosting:} Boosting is an ensemble strategy that learns from previous predictor errors in order to improve future predictions. The method incorporates many weak base learners into a single strong learner, greatly enhancing model predictability. Boosting operates by positioning slow learners in a sequence such that they can learn from the next learner in the sequence, resulting in more accurate predictive models.
        \item \textit{Stacking:} Another ensemble technique is stacking, which is also known as stacked generalization. This method operates by allowing a training algorithm to combine the predictions of several other similar learning algorithms. Regression, density estimations, distance learning, and classifications have all used stacking efficiently. 
    \end{enumerate}

\section{Proposed Method}

Before giving proper explanation of our proposed method, we want to first discuss the main methodology behind knowledge distillation and ensemble learning. 
\subsection{Distillation Theory} Neural networks typically produce class probabilities by using a “softmax” output layer that converts the logit, ${z_i}$, computed for each class into a probability, $q_i$, by comparing ${z_i}$ with the other logits.
\begin{center}
    ${q_i=\frac{e^{(z_i/T)}}{\sum_j e^{(z_j/T)}}}$
\end{center}
where $T$ is a temperature that is normally set to 1. As $T$ is set to a higher value, the probability distribution over classes becomes softer. In the most basic form of distillation, knowledge is transferred to the distilled model by training it on a transfer set and using a soft target distribution for each case in the transfer set, which is produced by using a cumbersome model with a high softmax temperature.  There  is  theoretical  study  on  how  the dark knowledge helps the student to learn finer structure from the big model, the teacher model. The low temperature models are usually good at  hard  predictions  and  we  lose  the  \textit{dark knowledge}. That's why, the main objective of distillation is to transfer the \textit{dark knowledge} from teacher to student.

\subsection{Importance of teaching assistant}
    In a traditional knowledge distillation scenario distillation is done from a teacher with large number of parameters which is directly
    proportional to the network size (here network size means number of convolutional layers in the
    CNN) to a small fixed student network (a convolutional neural network
    with 2 layers). It is done is such way that the student can retain the better accuracy of teacher because of its size. Here, a teacher
    model with 6, 8 10 layers are considered to train the student with 2 layers model on CIFAR-10 dataset. The
    interesting point that was observed is, student accuracy reduces with the increase in size of the teacher
    model even though the accuracy of that teacher model increases as shown in fig 3 where $n$ defines the number of hidden layers in the network architecture of the teacher model.
    \begin{figure}[htp]
    \centering
    \includegraphics[scale=0.6]{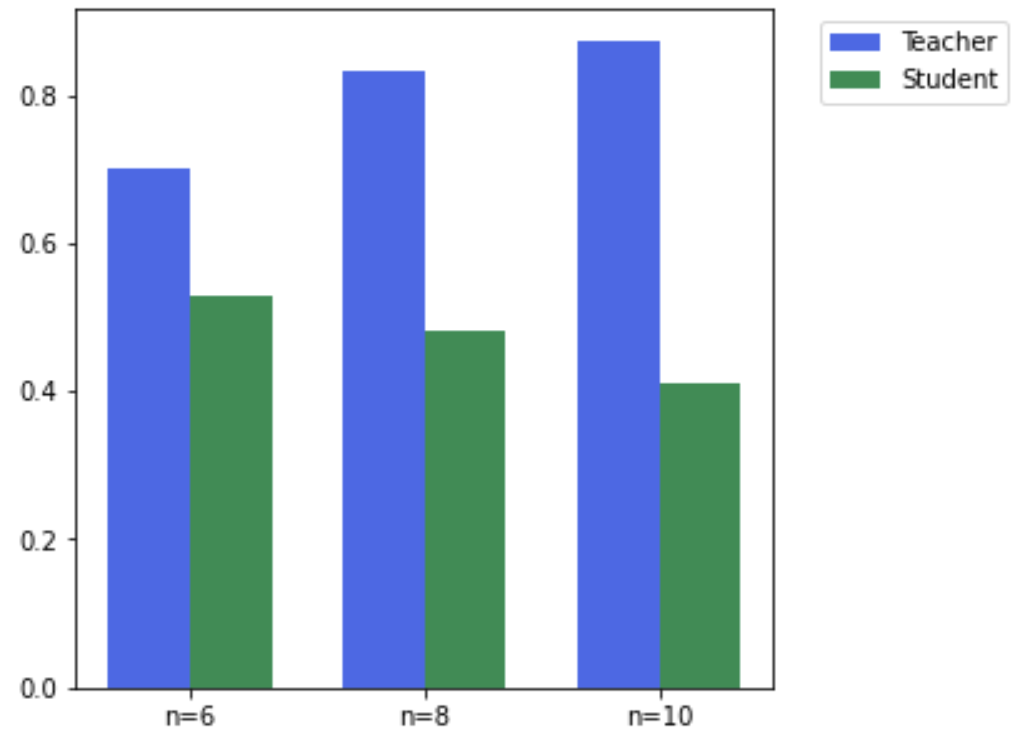}
    \caption{Performance of student model with teacher}
    \label{fig:n=3}
    \end{figure}
    \newline
    Few factors that actually that explains this drop in accuracy of student model are:
    \begin{enumerate}
        \item Teacher model is getting more complex that student does not have capacity to learn from the teacher model.
        \item Teacher’s certainty about data increases, thus making its logits (soft targets) less soft. This weakens the knowledge transfer which is done via matching the soft targets.
    \end{enumerate}
    \par
    To counter this problem an intermediate network is necessary which is having size between the size of teacher
    and student. That intermediate network gets distilled from teacher and distills its own knowledge to student. Here TAKD (teaching assistant Knowledge Distillation) model counter the problem 2 as its size is nearer to teacher and also it retains the accuracy of the teacher model. Student retains
    the teaching assistant accuracy as its size is nearer to the latter.

\subsection{Ensemble Model} Ensemble techniques are meta-algorithms that incorporate several machine learning strategies into a single predictive model to reduce variance, bias, and boost predictions (stacking). Neural network models are a nonlinear method. They can learn dynamic nonlinear interactions in data in this way. Their versatility is limited by their sensitivity to initial conditions, both in terms of initial random weights and statistical noise in the training dataset. Because of the stochastic nature of the learning algorithm, a neural network model can learn a slightly (or dramatically) different version of the mapping function from inputs to outputs each time it is trained, resulting in different results on the training and holdout datasets.
\par
As a result, a neural network may be thought of as a system of low bias and high variance. Even when trained on large datasets to satisfy the high variance, having any variance in a final model that is intended to be used to make predictions can be frustrating. A solution to the high variance of neural networks is to train multiple models and combine their predictions and the idea is to combine the predictions from multiple good but different models. There are few techniques to ensemble neural networks, and two of them are simple model averaging and extension of the model averaging called weighted model averaging.

    \subsection{Simple Model Averaging} The most basic method of ensemble learning is model averaging. It does just as the name implies: multiple models are trained on the same dataset, and we combine the results from all of them during prediction. The most popular approach for combining predictions in classification is to take votes from each model. For regression problems, we take the mean of each model's predictions. In general, the function we use to combine the predictions is determined by the cost function that is unique to the problem.
    \begin{figure}[htp]
    \centering
    \includegraphics[scale=0.25]{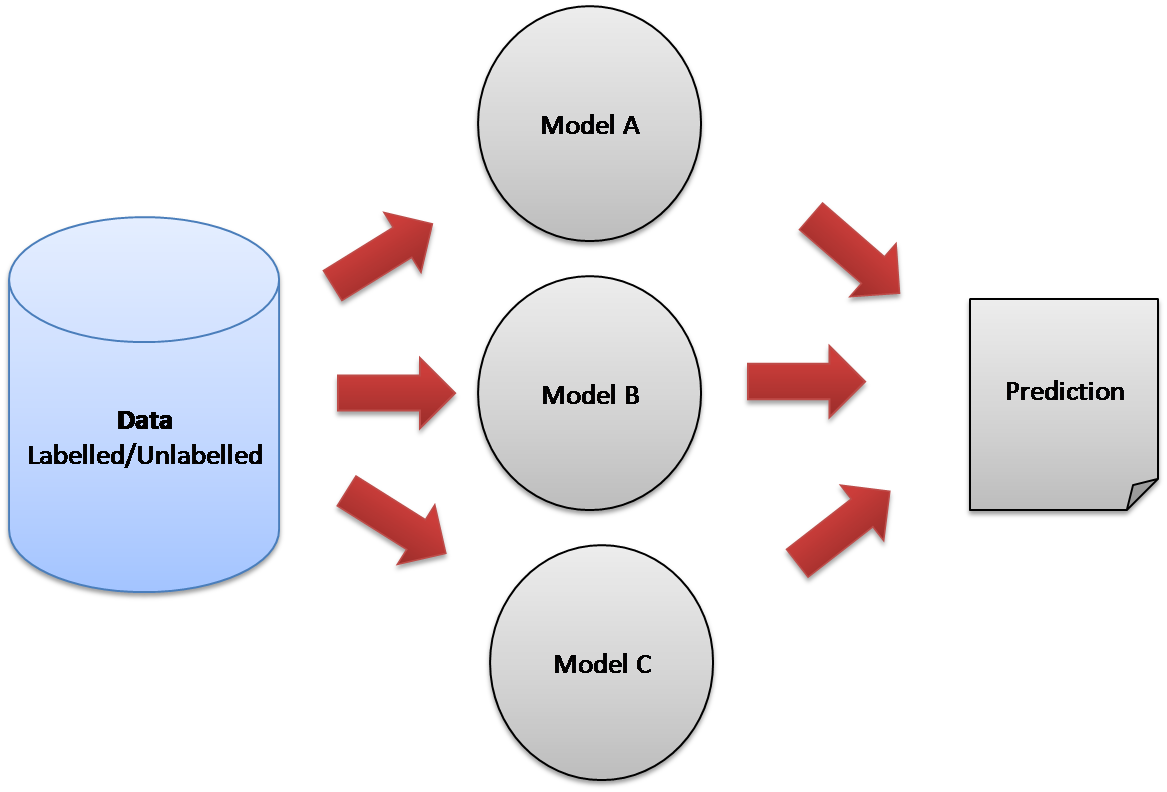}
    \caption{Simple model averaging}
    \label{fig:n=3}
    \end{figure}
    
    \subsubsection{Weighted Ensemble Model} Since each ensemble member will contribute equally to predictions, simple model averaging ensembles are constrained. Weighted average ensembles, on the other hand, cause each ensemble member's contribution to a prediction to be weighted proportionally to the member's confidence in results on the dataset. The model weights are small positive values and sum of all the weights equals one, allowing the weight indicates the percentage of expected performance of the network. Uniform values for the weights ($1/k$, where $k$ is number of ensemble members) means that weighted average ensemble acts as a simple averaging ensemble.
    \\
    \subsubsection{Our Proposed Method}
    In our experiment, we have used optimization techniques for searching the weights. The motivation behind using optimization process is to initiate a search process where instead of sampling the space of possible solutions randomly, it uses any available information to make the next step in search like toward a set of weights that has lower error. For our experiment, we have used SciPy library which provides many outstanding optimization algorithms. The one particular function we have used is "differential evaluation" method. This function provides stochastic global search algorithms that works for function optimization with continuous inputs. This function requires that function is specified to evaluate a set of weights and return a score to be minimized. It can be minimized using classification error(1- accuracy). This optimization makes use of loss function, condition for weights, and arguments search. Here, the condition is that each member’s weight should be between 0 and 1, their sum should be 1 and search arguments consists of teaching assistants, input values and actual output values. The loss functions return the evaluation of ensemble predictions of teaching assistants using weights. These weights are adjusted based on some step value which results in better accuracy.

     \begin{figure}[htp]
    \centering
    \includegraphics[scale=0.4]{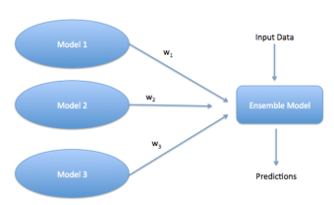}
    \caption{Weighted model averaging}
    \label{fig:n=3}
    \end{figure}

\section{Experiment Setup}
\par A set of experimentation were performed using the methodologies proposed such as knowledge distillation, teaching assistant models and ensemble of networks.

\subsection{Datasets}

The experiments are performed on CIFAR-10 [18], CIFAR-100 and MNIST datasets. The CIFAR-10 dataset contains 60,000 32 $\times$ 32 color images divided into ten groups, each with 6000 images. There are 50,000 images for teaching and 10,000 images for testing. The CIFAR-100 dataset contains 50,000 marked examples (500 per class) for training and 10000 unlabelled examples (100 per class) for testing, all of which are 32$\times$32 color images. MNIST dataset consists of 60,000 small square 28$\times$28 pixel grayscale images of handwritten single digits between 0 and 9, there are 60,000 examples in the training dataset and 10,000 in the test dataset.

\subsection{Software Implementation}
Keras is used to implement the neural network architecture and in preprocessing stage, both training and text data are normalized to ones with zero mean and standard deviation of 0.5. For optimization, stochastic gradient descent with Nesterov momentum of 0.9 and learning rate of 0.001 for 100 epochs. A suitable temperature value is suited for distilling the knowledge from teacher to teaching assistant to retain the dark knowledge of soft targets.

\subsection{Network Architecture}
For this experiment, a teacher model with size 10(number of layers) and a fixed student model of size 2 are considered. Five intermediate models, teaching assistants with size 8, 7, 6, 5 and 4 (between 2 and 10) is considered. The network architecture consists of convolutional cells followed by max pooling and ended with fully connected layer. The number of convolutional cells is considered as the network size. Output layer consists of softmax which means that the model will predict a vector with three elements with the probability that the sample belongs to each of the classes.

\section{Experiment Result}

\subsection{Traditional Knowledge Distillation Results}
Before diving into the comparison between our proposed method and traditional knowledge distillation we will first discuss the regular classification performance of our student model. 
Standalone performance of the student model is tabulated as table I. A visual representation of the result has been shown in fig. 9. 
\begin{table}[htp]
\caption{standalone student performance}
\centering
\begin{tabular}{|c|c|c|c|c|} \hline
    Model & Dataset & \multicolumn{1}{|p{2cm}|} {\centering Standalone student Accuracy}\\  \hline \newline 
    CNN  & CIFAR-10 & 48.01   \\
    CNN  & CIFAR-100 & 40.18  \\
    CNN  & MNIST & 85.41   \\
    \hline
\end{tabular}
\end{table}
   \begin{figure}[htp]
    \centering
    \includegraphics[scale=0.6]{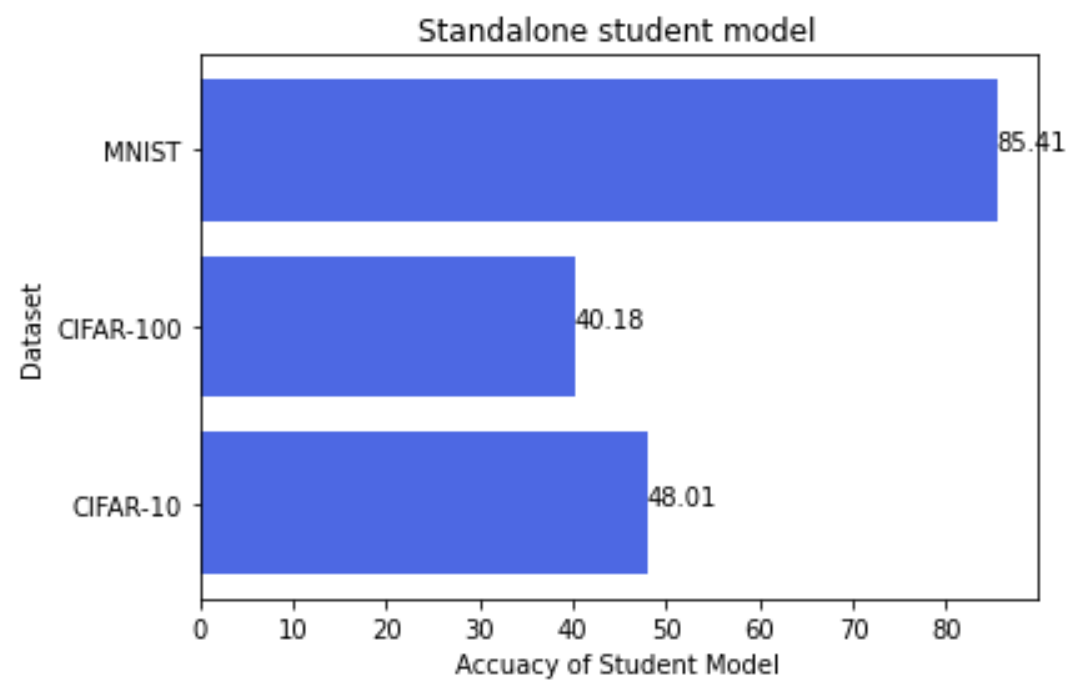}
    \caption{Accuracy of standalone student model }
    \label{fig:n=3}
    \end{figure}
    \\
Our next step is to implement the traditional knowledge distillation architecture. Using the proposed teacher and student architecture, base line knowledge distillation is implemented on the proposed datasets. The results are formulated in table II and fig 10. Now, if we compare the fig. 9 and fig. 10 we can clearly see the improvement of the student model using traditional knowledge distillation which was expected. \\
\begin{table}[htp]
\caption{student performance with base line knowledge distillation}
\centering
\begin{tabular}{|c|c|c|c|c|} \hline
    Model & Dataset & \multicolumn{1}{|p{2cm}|}{\centering Traditional Knowledge Distillation(Student Accuracy)}\\  \hline \newline 
    CNN  & CIFAR-10 & 49.32   \\
    CNN  & CIFAR-100 & 41.21  \\
    CNN  & MNIST & 86.59   \\
    \hline
\end{tabular}
\end{table}
 \begin{figure}[htp]
    \centering
    \includegraphics[scale=0.6]{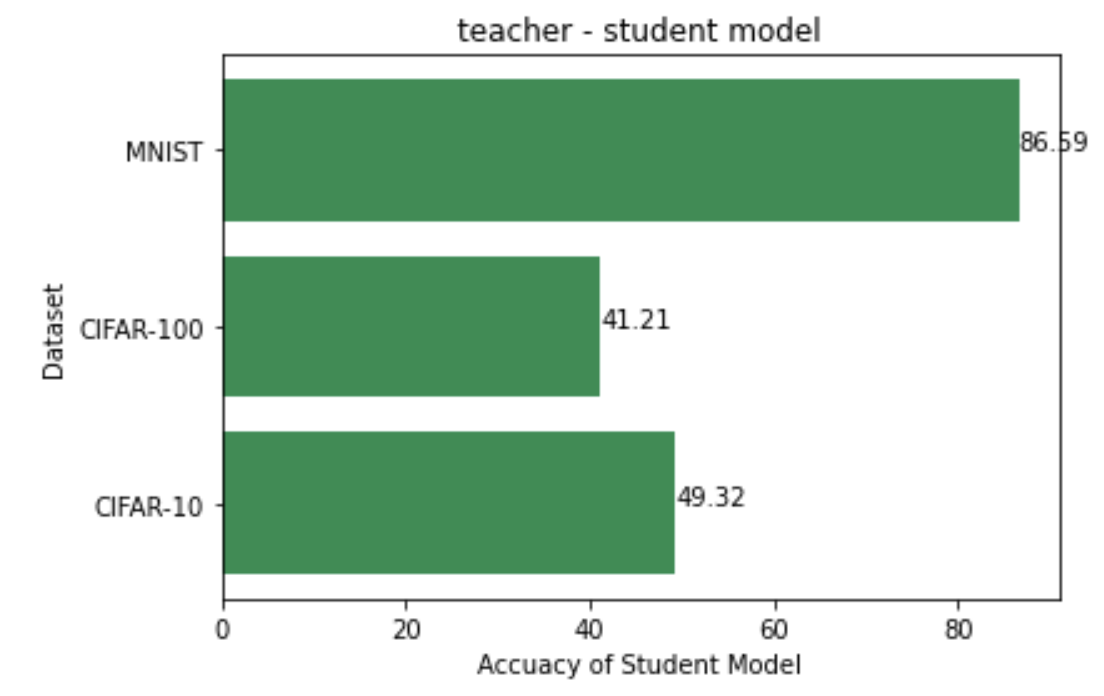}
    \caption{Accuracy of teacher-student model }
    \label{fig:n=3}
    \end{figure}
In the next phase our target was to use a single teaching assistant model and analyzing the result compare to the previous two results. 
Student performance is measured when teaching assistant model is used in between teacher and student models. For this experimentation, teaching assistant of size of 6 is considered and experimentation is performed on the proposed datasets. Results are tabulated in table III and accuracy can be seen in fig 11. From the results it is evident that student model performance can be increased by having a teaching assistant model in between teacher and student model.
\begin{table}[htp]
\caption{student performance with teaching assistant model}
\centering
\begin{tabular}{|c|c|c|c|c|} \hline
    Model & Dataset & \multicolumn{1}{|p{2cm}|}{\centering Single teaching assistant Model(Student Accuracy)}\\  \hline \newline 
    CNN  & CIFAR-10 & 50.21   \\
    CNN  & CIFAR-100 & 42.08  \\
    CNN  & MNIST & 87.73  \\
    \hline
\end{tabular}
\end{table}

 \begin{figure}[htp]
    \centering
    \includegraphics[scale=0.6]{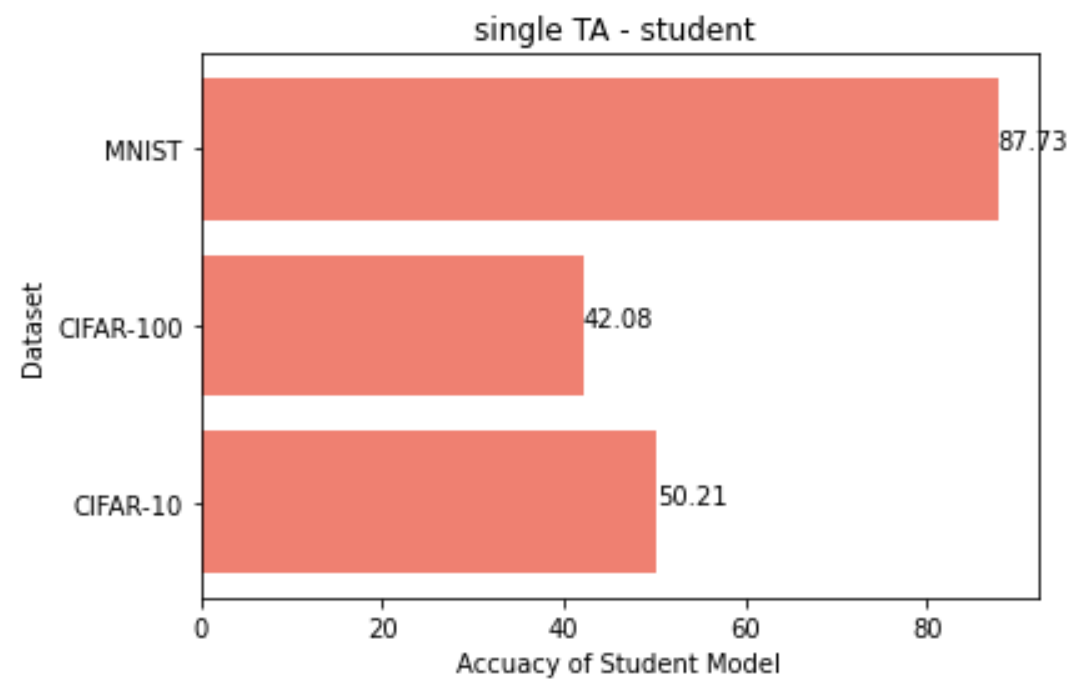}
    \caption{Accuracy of student model with teaching assistant }
    \label{fig:n=3}
    \end{figure}

\subsection{Our Proposed Model Results}

In this section, we will discuss on our proposed method which was ensemble of teaching assistant models and compare the performance in terms of student model accuracy. We have conducted two method of ensemble learning here. The simple model averaging and weighted model averaging. 
The results of proposed method where ensemble of teaching assistant models which have the distilled knowledge from same teacher is used to train the student model are shown in fig 12 and formulated in table IV. As discussed on section III, we have used differential evaluation optimization method to calculate the weights. Optimized weights for the teaching assistant models are obtained as follows: \\
\newline
for CIFAR-10 dataset: [0.2307629, 0.1891423, 0.2239231, 0.1891436, 0.1670281]   

for CIFAR-100 dataset: [0.2291469, 0.1747312, 0.2361738, 0.1943215, 0.1656266]   

for MNIST dataset: [0.2136427, 0.2061938, 0.2271425, 0.1749623, 0.1780587]  

\begin{table}[htp]
\caption{student performance trained with ensemble of TA networks}
\centering
\begin{tabular}{|c|c|c|c|} \hline
    Model & Dataset & \multicolumn{1}{|p{2cm}|}{\centering Ensemble network
    (Simple Model Averaging)}& \multicolumn{1}{|p{2cm}|}{\centering Ensemble network (Weighted Model Averaging)} \\  \hline \newline 
    CNN  & CIFAR-10 & 51.34 & 52.22  \\
    CNN  & CIFAR-100 & 43.36 & 44.79  \\
    CNN  & MNIST & 88.34 & 89.61  \\
    \hline
\end{tabular}
\end{table}

\begin{figure}[htp]
    \centering
    \subfloat[\centering simple model averaging]{{\includegraphics[width=8cm,height=8.05cm,keepaspectratio]{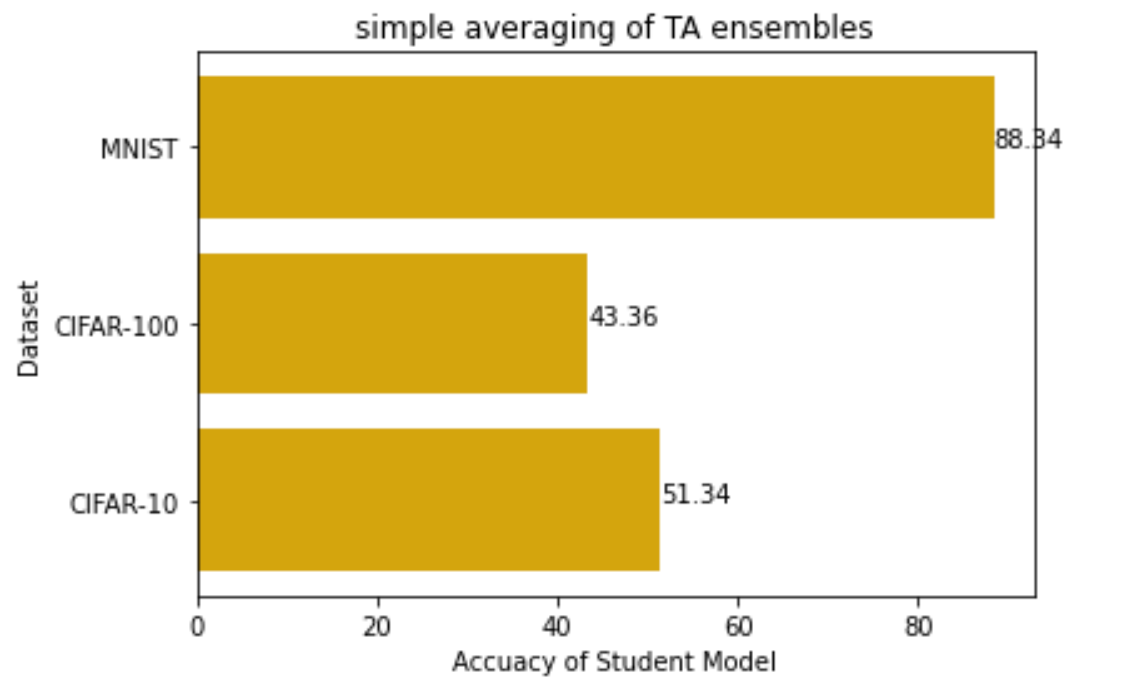} }}
    \qquad
    \subfloat[\centering weighted model averaging]{{\includegraphics[width=8cm,height=8.05cm,keepaspectratio]{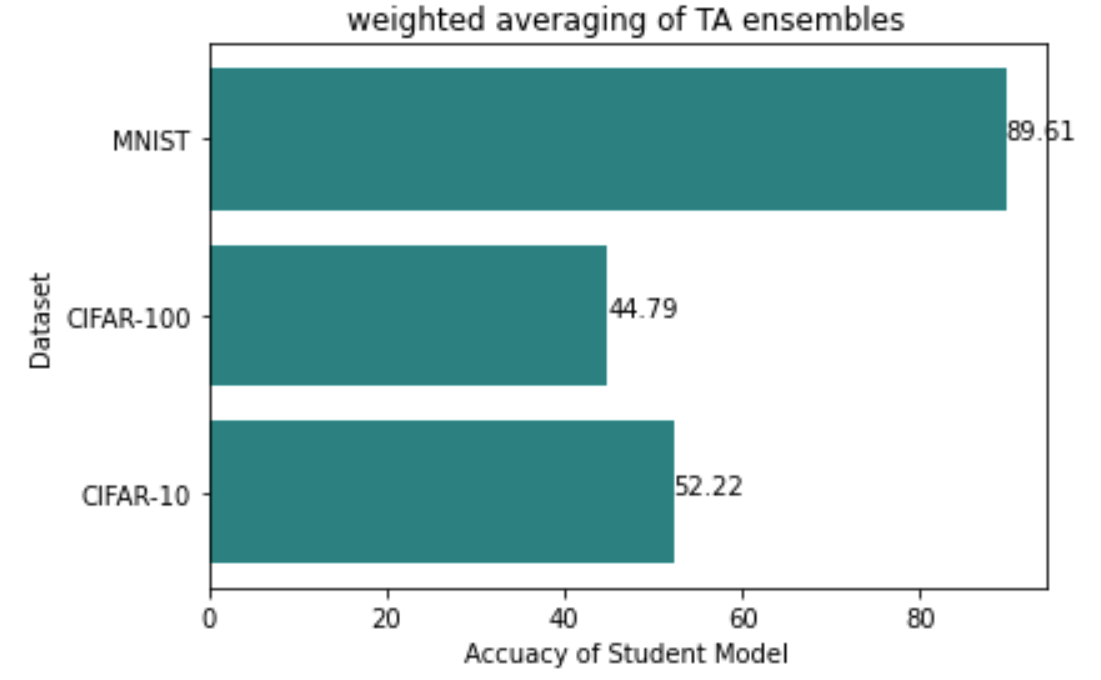} }}
    \caption{ student performance with ensemble of TAs}
    \label{fig:example}%
\end{figure}
\par 

\section{Result Analysis}

Now we are going to analyze our results corresponding to each dataset. We will compare the overall result in terms of performance improvement of the student model becuase that is our main goal to improve accuracy of the small model. \\
For CIFAR-10, simple model averaging achieved 3.33\% improvement over standalone student model and 2.02\% from traditional knowledge distillation method. Weighted model averaging got 4.21\% improvement over standalone student model and 2.9\% improvement over traditional knowledge distillation.
For CIFAR-100, simple model averaging achieved 3.18\% improvement over standalone student model and 2.15\% from traditional knowledge distillation method. Weighted model averaging got 4.61\% improvement over standalone student model and 3.58\% improvement over traditional knowledge distillation.
For MNIST dataset, simple model averaging achieved 2.93\% improvement over standalone student model and 1.75\% from traditional knowledge distillation method. Weighted model averaging got 4.2\% improvement over standalone student model and 3.02\% improvement over traditional knowledge distillation.
So, in conclusion we can draw this outline that "Weighted Ensemble of Teaching Assistant model $>$ Simple Averaging Ensemble Teaching Assistant model $>$ Traditional Knowledge distillation". Also, from the results we can clearly see the more the dataset is complex, the more performance improvement in terms of accuracy can be achieved using our proposed method. Fig 13, 14 and 15 is presented as visual re-presentation of those improvements.
 \begin{figure}[htp]
    \centering
    \includegraphics[scale=0.6]{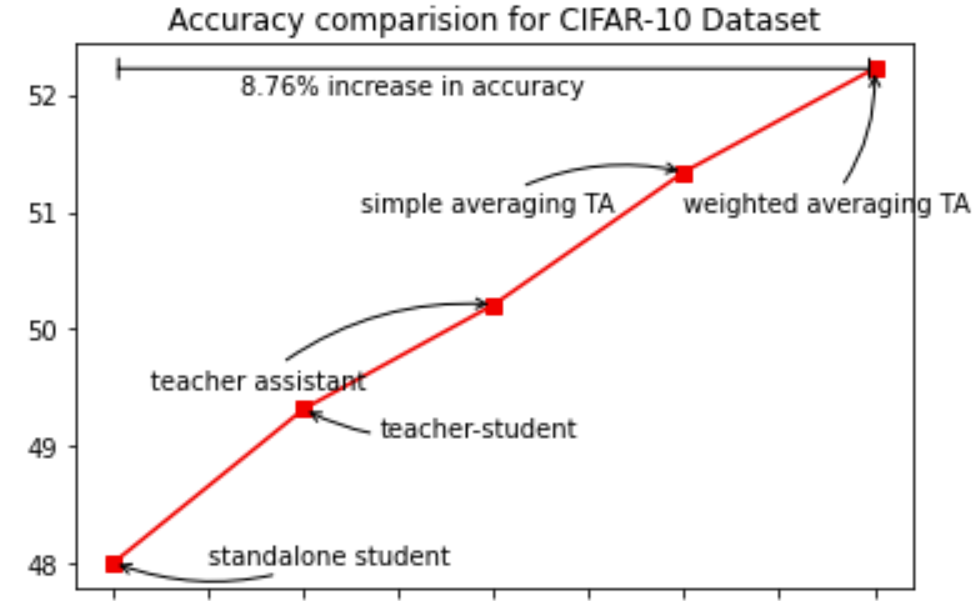}
    \caption{Performance of student model on CIFAR-10 }
    \label{fig:n=3}
    \end{figure}
     \begin{figure}[htp]
    \centering
    \includegraphics[scale=0.6]{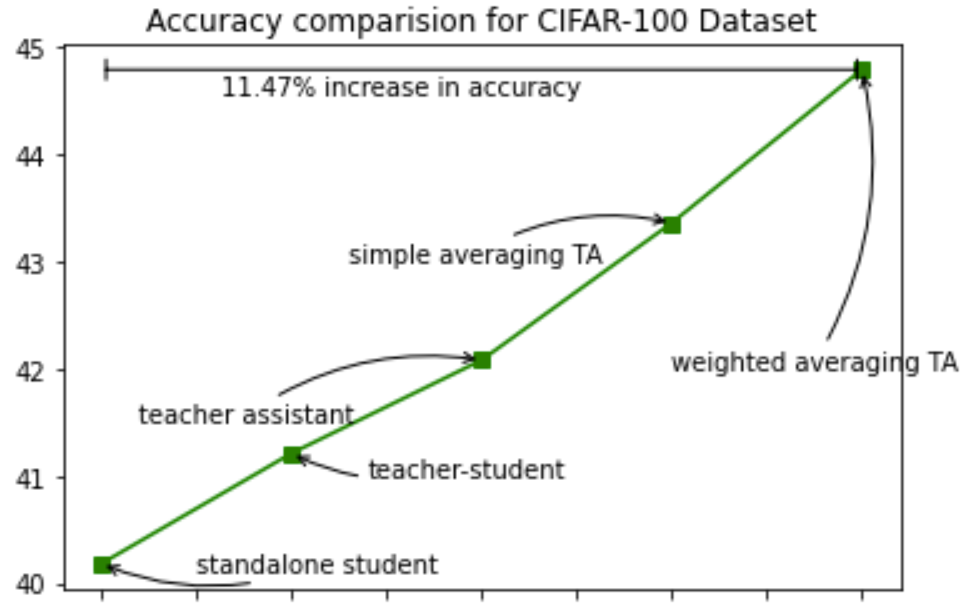}
    \caption{Performance of student model on CIFAR-100 }
    \label{fig:n=3}
    \end{figure}
     \begin{figure}[htp]
    \centering
    \includegraphics[scale=0.6]{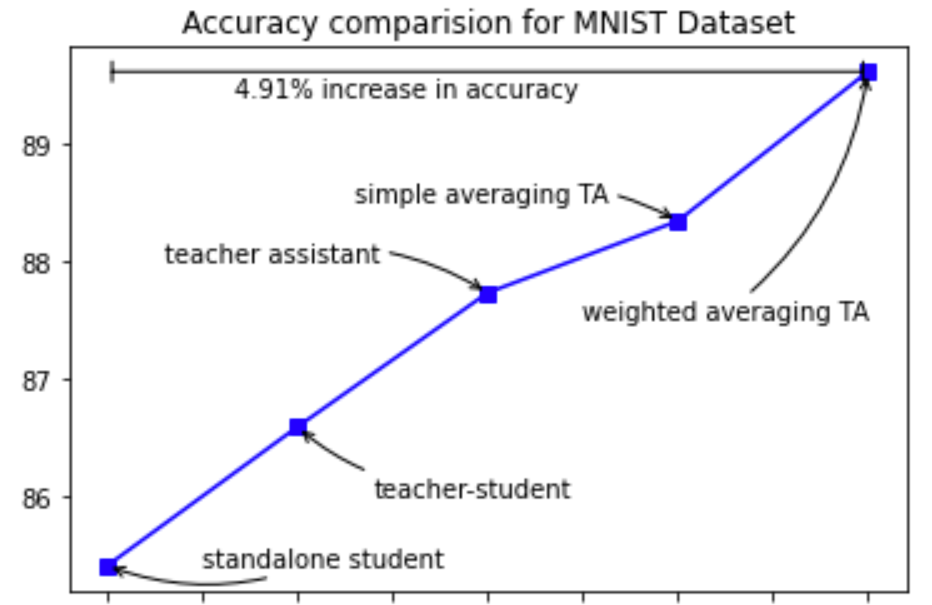}
    \caption{Performance of student model on MNIST }
    \label{fig:n=3}
    \end{figure}

\section{Conclusion and Discussions}

In this study, it is found that, student model performance can be actually improved by having ensemble of teaching assistant models in between teaching assistant model. And also, student trained by ensemble of TA performed moderately well than having a single TA. Performance of student model can be further increased with the implemented of weighted ensemble of teaching assistant model. Future research opportunities lies on calculating the optimal number of TA to gain performance improvement. Also variable temperature value for different teaching assistant can be good opportunity for future research.

\vspace{12pt}
LAO2112015E010E91
\end{document}